# Developing Robot Driver Etiquette Based on Naturalistic Human Driving Behavior

Xianan Huang, Songan Zhang, Huei Peng

*Abstract*—Automated vehicles can change the society by improved safety, mobility and fuel efficiency. However, due to the higher cost and change in business model, over the coming decades, the highly automated vehicles likely will continue to interact with many human-driven vehicles. In the past, the control/design of the highly automated (robotic) vehicles mainly considers safety and efficiency but failed to address the "driving culture" of surrounding human-driven vehicles. Thus, the robotic vehicles may demonstrate behaviors very different from other vehicles. We study this "driving etiquette" problem in this paper. As the first step, we report the key behavior parameters of human driven vehicles derived from a large naturalistic driving database. The results can be used to guide future algorithm design of highly automated vehicles or to develop realistic human-driven vehicle behavior model in simulations.

## I. INTRODUCTION

Automated vehicles can significantly change the future of ground mobility by reducing crashes, congestion, and fuel consumption. In addition, business model and cost/availability of mobility-on-demand service may also change when driverless vehicles become available. Mobility may be more accessible to the elderly and physically challenged population [1]. However, due to the cost differential, it is likely driverless vehicles will take a while to reach high market penetration [2]. In the next few decades, these robotic vehicles will operate in an environment interacting with many human-driven vehicles. According to reports from the California Department of Motor Vehicles (DMV) regarding autonomous vehicle on-road testing, most accidents involving driverless vehicles are caused by the surrounding human drivers [3]. After examining the crash rate of Waymo and Cruise Automation test fleets released by the California DMV, it becomes obvious that these driverless vehicles may be partially responsible for these crashes, even when the crashes are largely the responsibility of the other (human-driven) vehicle. The crash report of the Waymo fleet, for example, shows that they were crashed into by other vehicles much more often in 2015-2016 (13), compared with the crash rate of 2017 (3) [4], while the mileage is 636k miles in 2016 and 352k miles in 2017 in California. We hypothesize that it is not only important these vehicles do not crash into other vehicles, it is also important that they "merge into the local driving culture", and do not behave too differently from other (human-driven) vehicles, e.g., inappropriate driving speed, acceleration/deceleration, time headway, gap acceptance during lane change or left turn, etc. In other words, the robot vehicles must learn the "etiquette" of the local driving culture. In this paper, we report key parameters of human driving behaviors in three scenarios: free-flow driving, car-following, and lane-change/cut-in.

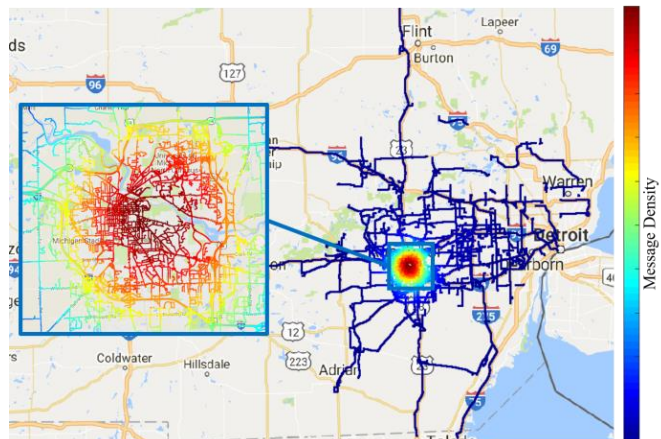

Fig. 1 Recorded vehicle location from Basic Safety Message (BSM) on May 1st, 2013 from the Safety Pilot Model Deployment (SPMD) database[5]

The robotic control of vehicle speed under free flow and car-following scenarios, e.g., applying to adaptive cruise control (ACC), has been studied extensively. Based on the longitudinal dynamics of the vehicle, sliding mode control [6], optimal dynamic back-stepping control [7] and adaptive control [8] have been used to design ACC. Also, car-following range or time headway policy and the influence on traffic were studied for homogeneous platoons [9][10] and mixed traffic [11]. It was found that proper headway policy can guarantee the string stability of platoons. Connected vehicle technologies such as Dedicated Short Range Communications (DSRC) [12] can be used to provide non-line-of-sight information such as platoon leader's acceleration, which enables cooperative adaptive cruise control [13]. With the knowledge of the motions of other vehicles, the CACC can stabilize a platoon which was string unstable [14]. However, a substantial portion of the work in the literature do not take human behaviors into consideration [15]. Related advanced driver assistance system (ADAS) work allow the driver to set the desired reference following distance and time headway [16] but the feedback control behavior may not be "human-like".

The lane change behavior has also been studied extensively. Hatipoglu et al. [17] designed an automated lane changing controller with a two-layer hierarchical architecture. Ammoun et al. [18] planned the desired lane changing trajectory with speed or acceleration constraints. In [19], Lee et al. proposed an integrated lane change driver model to control lane changing and lane following maneuvers. In [20], lane changes on curved roads were studied. In [21], lane change control under variable speed limits was shown to reduce travel time under various traffic density. However, in the literature whether these controlled lane change behaviors are compatible with human driving behavior were again largely not studied.

The behavior of human drivers has been collected in large-scale naturalistic field-operational-tests (N-FOTs). Driver characteristics such as time headway, range and range rate were studied [22], [23] and the behaviors were used to identify driver types [24]. Most of the studies in car-following focused on characterizing the control reference point of the human drivers, i.e. the desired car following distance and range rate, or capturing the influence on platoon dynamics [25]. For human lane change behavior, models usually are based on characteristics such as the range and gap at the initialization of a lane change [19]. Those models can be used to guide the design of autonomous vehicles. Moreover, in [26][27], the lateral acceleration during lane changes is captured. The information can guide the design of the lower-level controllers to ensure ride comfort. Finally, in [28], the duration of the lane change is analyzed. In this paper, we focus on the distribution of the initial range, initial Time to Collision (TTC), the maximum yaw rate of lane change vehicle, and duration of lane changes.

Considering both human driver behaviors already analyzed in the literature, as well as information we can extract from the data collected, we defined the key human driver behaviors to be analyzed, which are summarized in TABLE I.

TABLE I KEY BEHAVIOR VARIABLES USEFUL FOR THE DESIGN OF AUTOMATED VEHICLES

| |
| --- |
| Free Flow Speed |
| Time Headway in Car-Following |
| Range of Longitudinal Acceleration |
| Minimum Time Headway and Time to Collision (TTC) in Car-Following |
| Correlation between Acceleration and Range |
| Correlation between Acceleration and Range Rate |
| Maximum yaw rate during lane change |
| Range at the initiation of a lane change |
| Time to Collision (TTC) at the initiation of a lane change |
| Duration of lane changes |

The rest of this paper is organized as follows: Section 2 presents the naturalistic driving database used and the query criteria. Section 3 presents the results for three key scenarios: free-flow, car-following, and lane-changes. Conclusions and future work are given in Section 4.

## II. DATA DESCRIPTION

### A. Naturalistic Driving Database

The data used is from the Safety Pilot Model Deployment (SPMD) project lead by the University of Michigan Transportation Research Institute (UMTRI). SPMD data is collected from 2,800 passenger cars, trucks and buses equipped with DSRC devices to enable V2V and V2I communications and GPS to track vehicle motions. On the infrastructure side, there were 25 roadside equipment (RSE), 21 at signalized intersections, the remainder at curves and freeway locations.

The experiment has been running since August 2012 and has collected more than 5.6 TB of recorded Basic Safety Messages (BSM) including motion (speed, acceleration) and location (longitude, latitude) for all vehicles, Mobileye® and vehicle actuation (brake applied, traction control, etc.) information for some vehicles [29].

There are four types of vehicle equipment configurations, referred as Integrated Safety Device (ISD), Aftermarket Safety Device (ASD), Retrofit Safety Device (RSD), and Vehicle Awareness Device (VAD). The configurations are summarized in TABLE II. Among the 300 ASD vehicles, 98 were equipped with a Mobileye® camera, which records forward object, range, and lane tracking information.

### B. Sampled Dataset

#### 1) Car-Following

The key variables for the car-following scenario include the range between the host vehicle and the leading vehicle $R_L$, range rate $\dot{R}_L$, speeds of the host vehicle $v$ and the leading vehicle $v_L$, longitudinal accelerations of the host vehicle $a$ and the leading vehicle $a_L$, lane positions of the host vehicle $Y$ and the leading vehicle $Y_L$. We use data from 98 sedans equipped with Mobileye® which provides a) relative position to the leading vehicle (range) and b) lane tracking measures compared with the lane delineation both from the painted boundary lines and the road edge. The range measurements error is up to 10% at 90m and 5% at 45m [30]. To ensure consistency of the dataset, we apply the following query criteria:

- $R_L(t) \in [0.1 \text{ m}, 90 \text{ m}]$
- Latitude between 41.0º and 44.5º
- Longitude between -88.2º and -82.0º
- No cut-in vehicles between the two vehicles

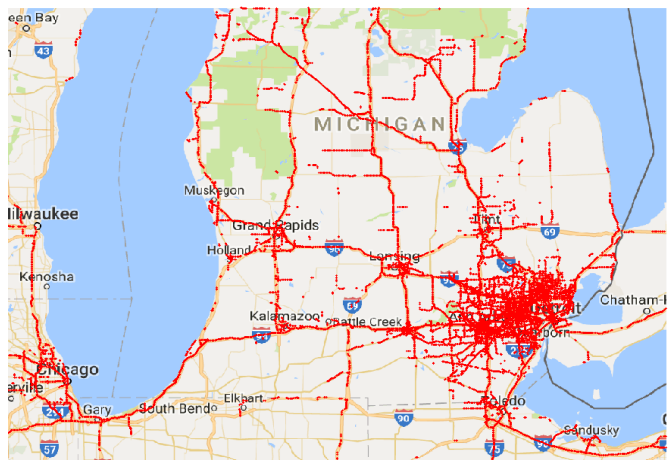

Fig. 2 Sampled car-following data location

TABLE II SPMD DSRC DEVICE SUMMARY

| Device | Tx | Rx | Weight Class | Quantity | Camera |
| --- | --- | --- | --- | --- | --- |
| ISD | Y | Y | Light | 67 | Y |
| VAD | Y | N | Light, Medium, Heavy Duty, Transit | 2450 | N |
| ASD | Y | Y | Light | 202 | N |
| | Y | Y | Light | 98 | Y |
| RSD | Y | Y | Heavy Duty, Transit | 19 | Y |

- No lane change by either vehicle
- Duration longer than 50s, $\dot{R} \in [-10\text{m/s}, 10\text{m/s}]$, vehicle speed larger than 10 m/s

With the defined criteria, 161,009 car-following events were identified: 85,656 on local roads and 75,353 on highways. The sampled car-following events are shown in Fig. 2.

### 2) Free-Flow Behavior

A Gaussian Mixture Model (GMM) based clustering algorithm is used to identify the free-flow condition from the data. The query criteria used for the trips are as follows:

- Trip duration longer than 10 minutes
- Trip length longer than 300 meters
- Trips inside the Ann Arbor area: latitude between 42.18° and 42.34°, and longitude between -83.85° and -83.55°

The results include 321,945 trips, which cover 3.7 million kilometers and more than 93,926 hours from 2,468 drivers. To match the trips to links (road sections), an algorithm developed by [31] is applied. The data covers 9,745 of the 11,506 road links in the Ann Arbor area.

### 3) Lane-Change

The lateral position of the POV reported from the Mobileye® camera is used to ensure the POV indeed cut-in in front of the host vehicle. As shown in Figure 3, the key lane change variables include the initial range to the leading vehicle $R_{L0}$, initial time-to-collision $TTC_0$, initial vehicle speed of the host vehicle $v_0$, the maximum yaw rate $r_{max}$ during a lane change, and the duration of lane change $T$. The query criteria used for the lane change scenario are as follows:

- Host vehicle is not changing lane
- Leading vehicle's lateral distance $d_{lat}$ to the host vehicle change from $d_{lat}(t_1) > 3m$ to $d_{lat}(t_2) < 0.3m$

In total, 422,249 cut-in cases were obtained. In 179,401 (42.5%) cases, the leading vehicle change lane from left to right, and in 242,848 (57.5%) of the cases, the leading vehicle change lane from right to left. 332,283 (78.7%) cases happen on local roads, and 89,966 (21.3%) cases happen on highways.

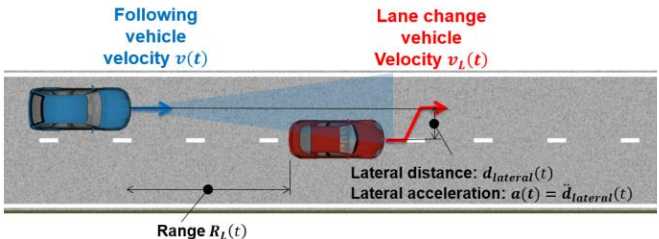

Fig. 3 Key variables extracted during a lane-change (cut-in) case

## III. RESULTS AND DISCUSSION

### A. Control Actions

#### 1) Longitudinal Acceleration and Deceleration

Longitudinal acceleration and deceleration characterize how decisive a vehicle is, and is an important behavior we study. On local roads, the distribution has a longer tail compared with that on highways. The longitudinal acceleration distribution of a selected driver is shown in Fig. 4. The distribution is asymmetric due to the difference in the powertrain acceleration and deceleration capabilities. In the following, we refer to deceleration as acceleration with brake applied, and acceleration as acceleration with throttle applied.

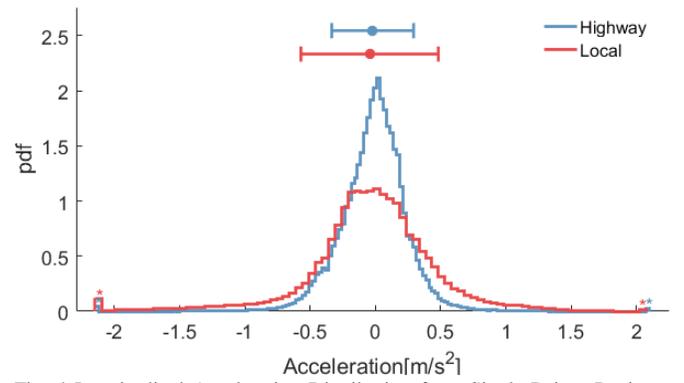

Fig. 4 Longitudinal Acceleration Distribution for a Single Driver During Car-Following for Highway and Local Driving

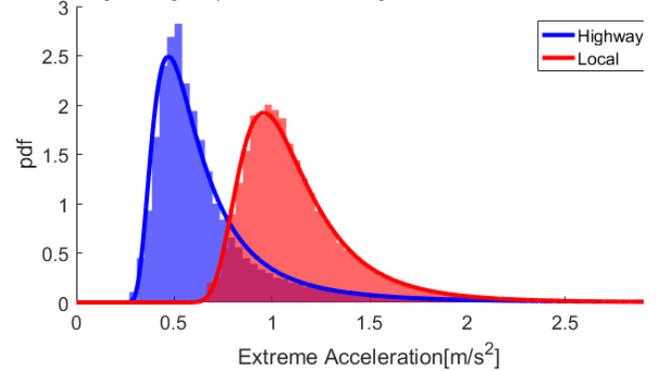

Fig. 5 Extreme acceleration distribution for all drivers

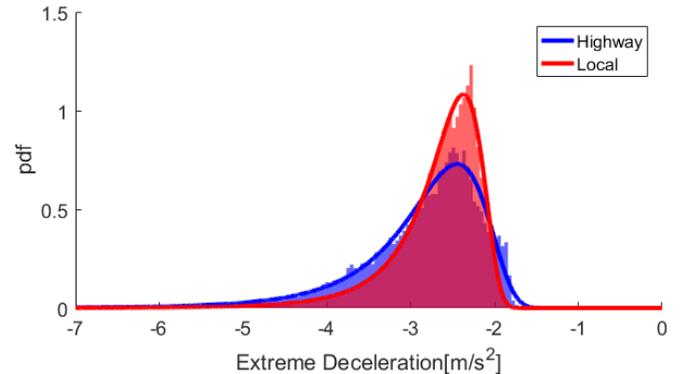

Fig. 6 Extreme deceleration distribution for all drivers

TABLE III
ACCELERATION AND DECELERATION LIMIT GEV
DISTRIBUTION PARAMETERS FOR HIGHWAY AND LOCAL CAR-FOLLOWING

| Scenario | | $k$ | $\sigma$ | $\mu$ |
|---|---|---|---|---|
| Highway | $a_{lim,a}$ | 0.3711 | 0.1628 | 0.5314 |
| | $-a_{lim,d}$ | 0.1669 | 0.4722 | 2.4461 |
| Local | $a_{lim,a}$ | 0.1426 | 0.1930 | 1.0457 |
| | $-a_{lim,d}$ | 0.1649 | 0.3289 | 2.3865 |

For each driver, we define deceleration stronger than 2.5% percentile as extreme deceleration and acceleration stronger than 97.5% percentile as extreme acceleration. The extreme acceleration $a_{lim,a}$ and deceleration $a_{lim,d}$ of all drivers are shown in Fig. 5 and Fig. 6, respectively. The distributions are fitted with a Generalized Extreme Value (GEV) distribution model. The parameters of the GEV distribution include shape parameter $k$, scale parameter $\sigma$ and location parameter $\mu$, the probability function is shown below

$$f(x|k,\mu,\sigma) =$$
$$\begin{cases} \dfrac{1}{\sigma}\exp\left(-\left(1+\dfrac{k(x-\mu)}{\sigma}\right)^{-\frac{1}{k}}\right)\left(1+\dfrac{k(x-\mu)}{\sigma}\right)^{-1-\frac{1}{k}} & k \neq 0 \\[2mm] \dfrac{1}{\sigma}\exp\left(-\exp\left(-\dfrac{x-\mu}{\sigma}\right)-\dfrac{x-\mu}{\sigma}\right) & k = 0 \end{cases}$$
(1)

The parameters are summarized in TABLE III. Human drivers have higher acceleration levels on local roads than on highways, with average acceleration limit 0.72 m/s² for highway and 1.19 m/s² for local roads. The mean deceleration limit for highway and local car-following are close, with -2.81 m/s² for highway and -2.64 m/s² for local roads. However, the tail for highway deceleration is longer than local driving.

### 2) Maximum Yaw rate During Lane Changes

To prevent a robotic vehicle executing a lane change too aggressively, it is important to learn human lane change maximum yaw rate $r_{max}$. In this section, the distributions of $r_{max}$ of local roads and highway are analyzed. The yaw rate of lane change vehicle is calculated from a Kalman Filter using the time series of lateral distance of the lane change vehicle. Assuming the initial yaw rate of each lane chage is zero, calculate the yaw rate time series using Kalman Filter derivation described in [32] with state variance matrix used in [33]. As can be seen from Fig. 7, the average maximum yaw rate of the local lane change (1.4 deg/s) is much higher than the highway lane change (0.6 deg/s) and the local lane change maximum yaw rate has a longer tail. This is due to the lower driving speeds on local roads.

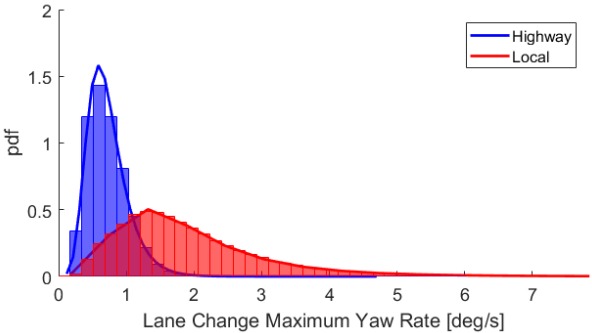

Fig. 7 Lane Change Maximum Yaw Rate Distribution

TABLE IV MAXIMUM YAW RATE GEV DISTRIBUTION PARAMETERS FOR HIGHWAY AND LOCAL LANE CHANGE

| Scenario | | $k$ | $\sigma$ | $\mu$ |
|---|---|---|---|---|
| Highway | $r_{max}$ | -0.0083 | 0.2325 | 0.5900 |
| Local | $r_{max}$ | 0.1525 | 0.7381 | 1.3953 |

### B. Free Flow Behavior

The free flow driving behavior was studied extensively in the literature [34]. Measurement data from roadside sensors show that the traffic flow demonstrates a multimodal behavior, which was commonly described by a three-phase traffic theory: free flow, synchronized flow, and wide-moving jam. The latter two phases are associated with congested traffic. Based on this theory, we use the Gaussian Mixture Model (GMM) [35] with 3 components to identify the free flow and congested behaviors.

$$f\left(x|\pi_{1,\ldots,k},\theta_{1,\ldots,k}\right) = \sum_{k=1}^{3}\pi_k f_k(x|\theta_k)$$
(2)

where $\pi_k$ is weighting parameters, and $f_k(x|\theta_k)$ is the multivariate normal probability density function of each component, $\theta_k$ is the collection of model parameter of each component, which includes mean and covariance matrix. The model assumes that the congestion status can be viewed as a discrete random variable, and the vehicle speed is a random variable conditional on the congestion status. Samples of local and highway speed models for one road section are shown in Fig. 8 and Fig. 9.

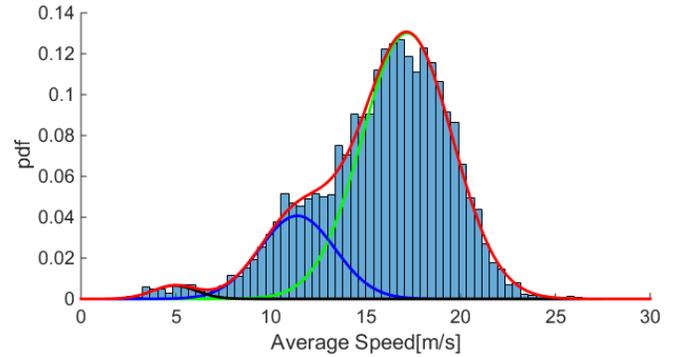

Fig. 8 Speed histogram and GMM fitting for one local road section with a speed limit at 17.88 m/s (40 mph)

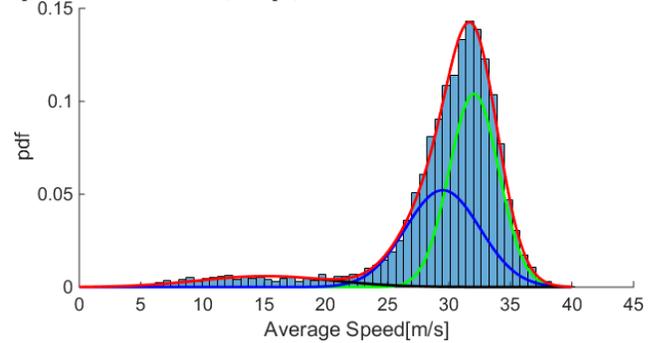

Fig. 9 Speed histogram and GMM fitting for one highway road section with a speed limit 31.29 m/s (70 mph)

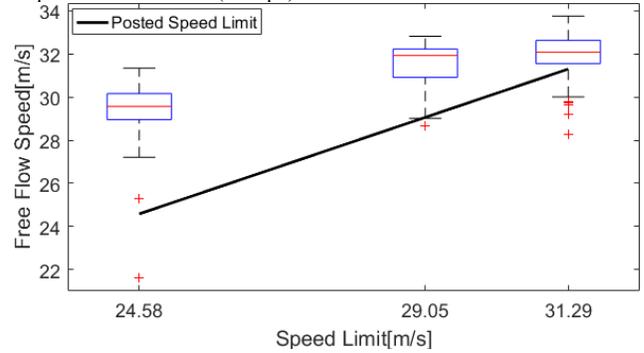

Fig. 10 Free Flow Speed vs. Posted Speed Limit for Highways

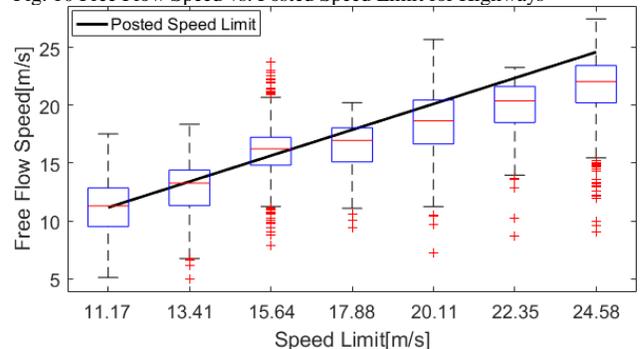

Fig. 11 Free Flow Speed vs. Posted Speed Limit for Local Roads

We use the component with the highest mean value for each link to estimate the free flow behavior. The measured free-flow speed compared with the posted speed limits are shown in Fig. 10 and Fig. 11, where the observed free flow speed of the road links vs. speed limits shown in a box plot and posted speed limit shown in a solid line. As shown in the figures, the observed free-flow speed is significantly higher than the posted speed limits on the highways. According to the Highway Capacity Manual [36], the base free-flow speed is estimated to be 2.2 m/s (5 mph) above the posted speed limit. However, as shown in Fig. 10, for highway links with lower speed limits, the HCM estimated base free-flow speeds are much lower than the measured values. This could pose a dilemma for robot drivers—if the robots are programmed to follow the speed limit, they will drive much slower than human driven vehicles, especially on highways with slower posted speed limit (e.g., 45 mph). For local roads, the mean free-flow speeds are very close to the posted speed limits, with a correlation coefficient of 0.99.

### C. Car-following Behavior

#### 1) Distance to the Lead Vehicle

The relative position from a host vehicle to the lead vehicle can be defined by the time headway, which is range divided by the speed of the host vehicle. The constant time headway policy is frequently used as a safe driving practice for human drivers and for Adaptive Cruise Control designs. Two key statistic parameters are the average time headway and minimum time headway. For human drivers, the lognormal function was found to fit their time headway distribution well [37]. The sample time headway distribution of a single driver for both highway and local car-following events are shown in Fig. 12. As shown in the histograms, the sampled driver tends to keep a longer time headway on local roads, and the variance is larger, compared with the behavior on highways.

To model the time headway distribution for the entire driver population, the mean time headway for each driver is calculated and plotted in Fig. 13. The distribution is fitted using a lognormal distribution function, and the parameters are summarized in TABLE V. The mean car-following time headway for highway driving is 1.42 s. Our highway results agree with previous studies such as [38] which concluded that car-following time headway for highway is between 1.3 s and 1.6 s, which correspond to 25% and 75% percentiles of our model. The 25% and 75% percentiles of local road sections are 1.77 s and 2.33 s. From the histograms, time headway for local roads is longer than that of highways, which has a median of 2.03 s and an average of 2.07 s.

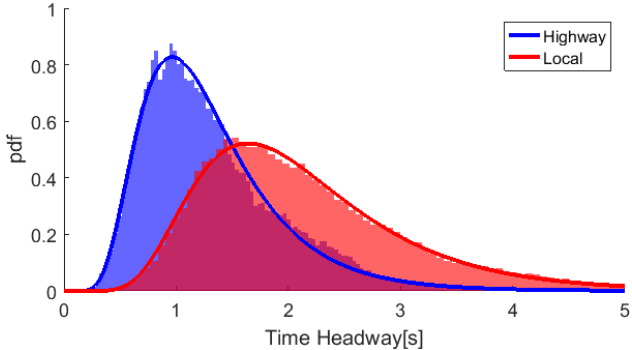

Fig. 12 Time Headway Distribution and Lognormal Model for Single Driver in Car-Following Scenario for Highway and Local Driving

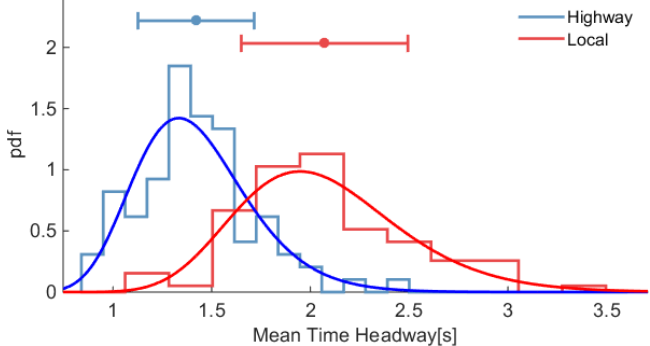

Fig. 13 Mean time headway distribution of all drivers for highway and local car-following events

TABLE V MEAN CAR-FOLLOWING TIME HEADWAY LOGNORMAL DISTRIBUTION PARAMETERS AND PERCENTILE

| Scenario | Mean [s] | Variance [s²] | Percentile [s] | | |
|---|---|---|---|---|---|
| | | | 25% | 50% | 75% |
| Highway | 1.42 | 0.08 | 1.21 | 1.39 | 1.60 |
| Local | 2.07 | 0.18 | 1.77 | 2.03 | 2.33 |

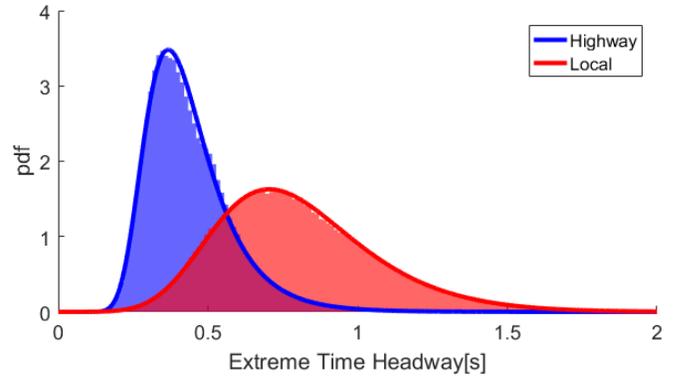

Fig. 14 Extreme time headway distribution for all drivers

TABLE VI TIME HEADWAY LIMIT GEV DISTRIBUTION PARAMETER FOR HIGHWAY AND LOCAL CAR-FOLLOWING

| Scenario | | $k$ | $\sigma$ | $\mu$ |
|---|---|---|---|---|
| Highway | $Th_{lim}$ | 0.0415 | 0.1058 | 0.3720 |
| Local | $Th_{lim}$ | -0.0737 | 0.2267 | 0.6880 |

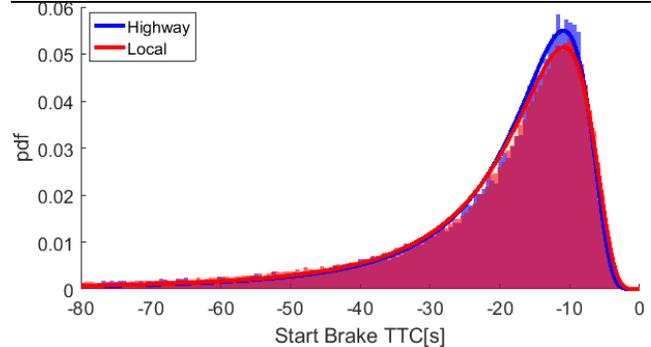

Fig. 15 Starting-to-brake TTC of highway and local car-following events

TABLE VII START-TO-BRAKE TTC GEV DISTRIBUTION PARAMETER FOR HIGHWAY AND LOCAL CAR-FOLLOWING

| Scenario | | $k$ | $\sigma$ | $\mu$ |
|---|---|---|---|---|
| Highway | $|TTC_{lim}|$ | 0.4006 | 7.1869 | 13.1760 |
| Local | $|TTC_{lim}|$ | 0.3989 | 7.6780 | 13.2650 |

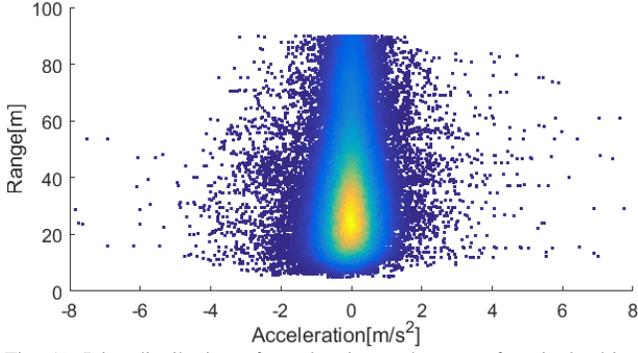

Fig. 16 Joint distribution of acceleration and range of a single driver highway car-following scenario

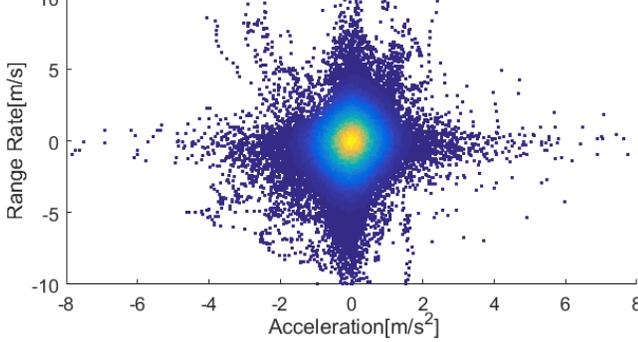

Fig. 17 Joint distribution of acceleration and range rate of a single driver highway car-following scenario

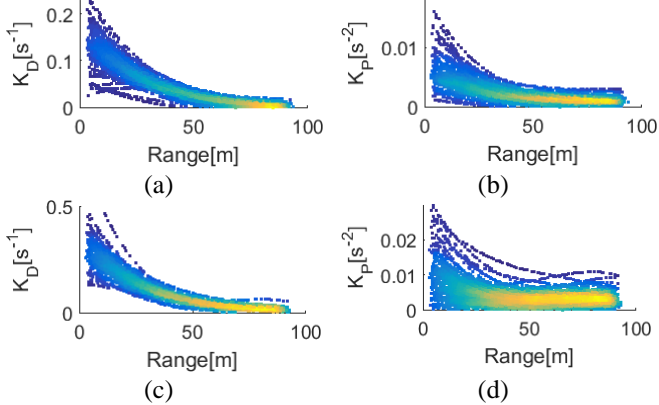

Fig. 18 Joint distribution of correlation and range for different scenarios: (a) $K_D$ for highway; (b) $K_P$ for highway; (c) $K_D$ for local; (d) $K_P$ for local

The minimum car-following distance is also calculated for all drivers. For each driver, the extreme time headway is defined as 2.5% percentile of the distribution of that driver. The extreme time headway of all drivers are shown in Fig. 14. The random variables are characterized with GEV distributions, and the parameters are summarized in TABLE VI. The extreme time headway on highways is 0.44 s, shorter than the 0.80 s for local roads. The standard deviation is 0.021 s on highways, lower than that that of the for local roads (0.071 s).

In addition to time headway, another variable commonly used to characterize driving is time to collision (TTC), which is defined as the ratio between range and the absolute value of range rate. Since the closing-in process is of interest, we only analysis the cases when range rate is negative. The "starting-to-brake TTC" is the TCC when the human drivers started to apply the brake, for both highway and local car-following events, are shown in Fig. 15. The distributions are again characterized with a GEV distribution. The model parameters are obtained from maximum likelihood estimation and are summarized in TABLE

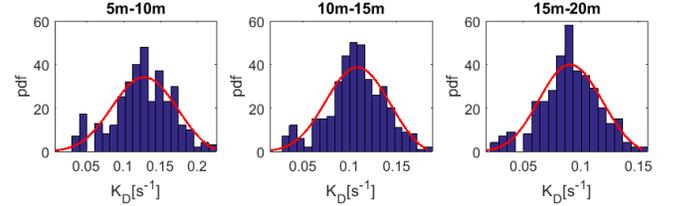

Fig. 19 Distribution of $K_D$ from 5 m to 20 m in highway car-following

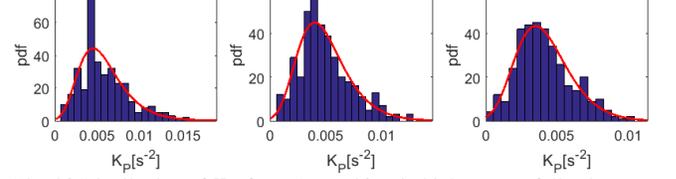

Fig. 20 Distribution of $K_P$ from 5 m to 20 m in highway car-following

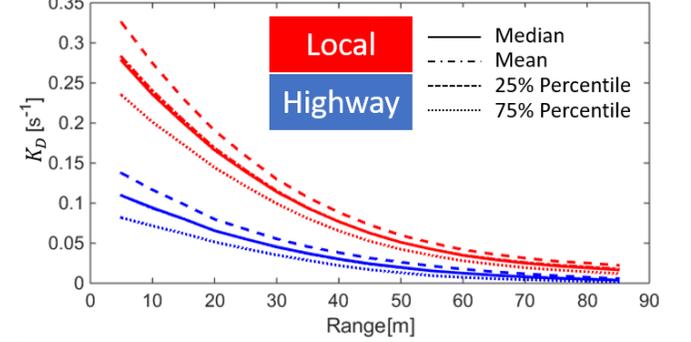

Fig. 21 Mean correlation between acceleration and range rate at different ranges for highway and local car-following scenarios

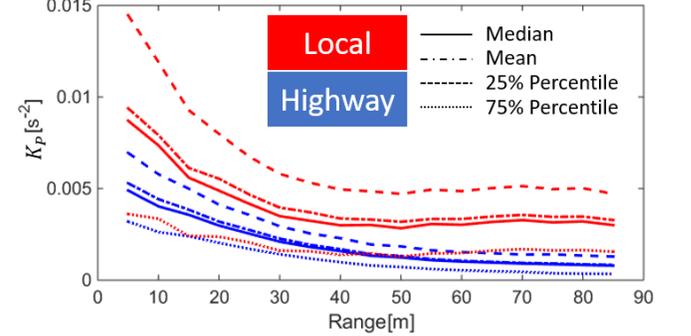

Fig. 22 Mean correlation between acceleration and range error at different ranges for highway and local car-following scenarios

VII. The results indicate that the starting-to-brake TTC for highway and local car-following cases are similar, with the average value at around 22 s, while the mode is at 12 seconds.

*2) Dynamic Response to the Lead Vehicle*

The dynamic response of human drivers to the lead vehicle can be understood by the correlation between acceleration (the control action) to the range and range rate (the vehicle states), e.g., following the driver model proposed by [37]. In this model, both correlations are modeled as a function of range $R_L$. The acceleration can be expressed as

$$a_d = K_D(R_L)\dot{R}_L + K_P(R_L)(R_L - Th_d \cdot v) \quad (3)$$

where $K_D$ is the control gain for the range rate, and $K_P$ is the control gain for the range, $Th_d$ is the desired time headway to the lead vehicle. The sample joint distributions for range, range rate and acceleration of a single driver are shown in Fig. 16 and Fig. 17. At longer range, the variance of acceleration decreases, indicating human drivers are less sensitive. The correlations are modeled as a 3rd order polynomials in range. The parameters are estimated using robust least square with a bisquare function as regularization weights. [39] With this algorithm, the parameter estimation is more robust against outliers. For each driver, the correlation polynomials are estimated, and then the results for all drivers are used to construct the model of correlation parameters. The joint distribution for the correlations and range are shown in Fig. 18. It can be seen that the drivers use higher feedback gains when they are closer to the lead vehicle. The GEV distribution is used to model the random variable to capture the asymmetricity. The distribution from 5 m to 20 m for highways are shown in Fig. 19 and Fig. 20 as an example.

With distribution parameters estimated for all car-following cases, the population mean of the correlation function and the percentiles are computed. The mean correlation and 25% and 75% percentiles at different car-following ranges are shown in Fig. 21 and Fig. 22.

### D. Lane-Change Behavior

#### 1) Range at Initiation of a Lane Change

The initial range of a lane change maneuver is an important parameter to characterize cut-in behaviors. Human drivers examine the available adjacent gaps to decide whether to change lane [40]. Therefore, it is crucial to understand how human drivers accept the gap for lane changes. As shown in [41], the initial range reciprocal is usually used in the models. In this section, the distribution of initial range reciprocal is fitted using GEV (see Eq.(1)). All the fitted parameters are shown in TABLE VIII. As shown in Fig. 23, the mean initial range of lane change on the highways (75.8m) is longer than the initial range of lane change on local roads (60.2m). Other percentile of range at initiation is also shown in TABLE VIII. An important observation is that if robot drivers are designed to be no more aggressive than the 90% lane change conducted by human drivers, then the shortest range at the initiation of lane changes is 17 meters on the highways and 22 meters for local roads.

#### 2) Initial Time to Collision (TTC) of Lane Change

As in [41], the initial TTC reciprocal is analyzed. Positive TTC represents cases when the following vehicle is catching up to the leading lane change vehicle. The higher initial TTC reciprocal, the more risky the lane change is. We use double exponential distribution to capture both negative and positive TTC, with the probability density function described as:

$$f(x) = \frac{\lambda}{2}\exp(-\lambda|x-\mu|) \quad (4)$$

where $\lambda$ is shape parameter, and $\mu$ is location parameter which indicate the mean value of dataset. And for positive initial TTC (dangerous) cases, the distribution is fitted by exponential distribution, with the probability density function described as:

$$f(x) = \frac{1}{\mu}\exp(-x/\mu) \quad (5)$$

where $\mu$ is the mean parameter.

The distribution is shown in Fig. 24 and Fig. 25 and model parameters are shown in TABLE IX and TABLE X for initial TTC and positive initial TTC respectively. The positive TTC percentile is also shown in TABLE X.

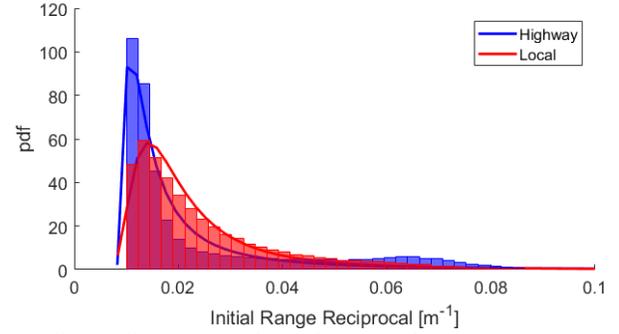

Fig. 23 Lane Change Initial Range Reciprocal Distribution

TABLE VIII  INITIAL RANGE RECIPROCAL GEV DISTRIBUTION PARAMETERS FOR HIGHWAY AND LOCAL LANE CHANGE

| Scenario | | $k$ | $\sigma$ | $\mu$ |
|---|---|---|---|---|
| Highway | $R_{L0}^{-1}$ | 0.8429 | 0.0049 | 0.0132 |
| Local | $R_{L0}^{-1}$ | 0.4495 | 0.0069 | 0.0166 |
| Percentile | 10% | 30% | 70% | 90% |
| Highway | 1/ 96.2 | 1/82.3 | 1/44.8 | 1/ 17.0 |
| Local | 1/85.7 | 1/66.5 | 1/36.8 | 1/ 21.8 |

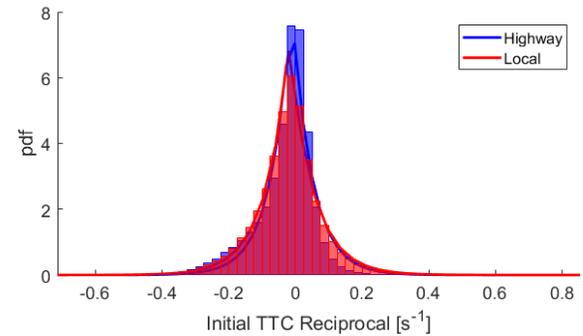

Fig. 24 Lane Change Initial TTC Reciprocal Distribution

TABLE IX  INITIAL TTC RECIPROCAL  DOUBLE EXPONENTIAL DISTRIBUTION PARAMETERS FOR HIGHWAY AND LOCAL LANE CHANGE

| Scenario | | $\lambda$ | $\mu$ |
|---|---|---|---|
| Highway | $TTC_0^{-1}$ | 16.5370 | -0.0120 |
| Local | $TTC_0^{-1}$ | 14.0112 | -0.0185 |

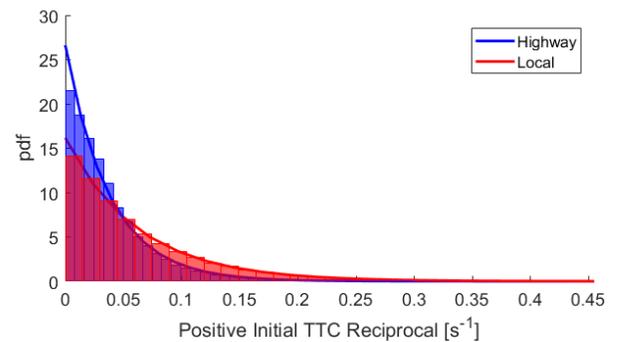

Fig. 25 Positive Initial TTC Reciprocal Distribution

TABLE X  POSITIVE INITIAL TTC RECIPROCAL DOUBLE EXPONENTIAL DISTRIBUTION PARAMETERS FOR HIGHWAY AND LOCAL LANE CHANGE

| Scenario | | $\mu$ |
|---|---|---|
| Highway | $+TTC_0^{-1}$ | 0.0376 |
| Local | $+TTC_0^{-1}$ | 0.0619 |
| Percentile | 10% | 30% | 70% | 90% |
| Highway | 1/219.7 | 1/68.0 | 1/22.5 | 1/12.1 |
| Local | 1/148.5 | 1/45.0 | 1/13.4 | 1/6.95 |

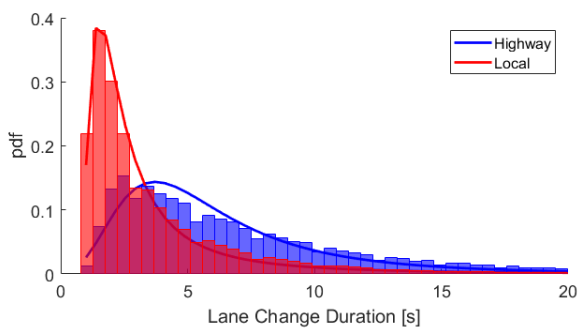

Fig. 26 Lane Change Duration Distribution

TABLE XI   DURATION GEV DISTRIBUTION PARAMETERS FOR HIGHWAY AND LOCAL LANE CHANGE

| Scenario | | k | σ | μ |
|---|---|---|---|---|
| Highway | T | 0.2675 | 2.6435 | 4.3243 |
| Local | T | 0.6585 | 1.1138 | 2.0052 |
| Percentile | 10% | 30% | 70% | 90% |
| Highway | 2.2 | 3.6 | 8.1 | 13.1 |
| Local | 1.2 | 1.8 | 4.0 | 7.8 |

### 3) Duration of Lane Change

In this section, the distribution of lane change duration $T$ is fitted by GEV. As shown in Fig. 26, the mean duration of lane change in highway (4.3s) is longer than the duration of lane change in local road (2.0s). The distribution parameters and duration percentile is shown in TABLE XI.

## IV. CONCLUSION AND FUTURE WORK

In this paper we present the key parameters of human driver behaviors in three scenarios: free-flow, car-following and lane-change, obtained from a naturalistic driving database.

Our results can be used to design control algorithm of automated vehicle so that it is compatible to local driving culture. The results can also be used to develop driving simulation software to simulate human-driven vehicles. Our next step includes developing automated vehicle based on the parameters and validate the functions in testing facilities such as Mcity [42].

## V. ACKNOWLEDGMENT

The authors would like to thank University of Michigan Transportation Research Institute (UMTRI) for making the data from the Safety Pilot Model Deployment (SPMD) project available.

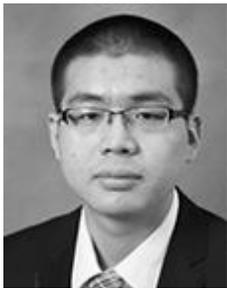
**Xianan Huang** received B.S. degree in mechanical engineering from Shanghai Jiaotong University and Purdue University in 2014. He is currently pursuing the Ph.D degree in mechanical engineering at University of Michigan, Ann Arbor.

From 2013 to 2014 he was an undergraduate researcher at Purdue University. Since 2014 he has been a graduate researcher at University of Michigan, Ann Arbor. His research interests include connected automated vehicle, intelligent transportation system, statistical learning and controls.

Mr. Huang's awards and honors include A-Class scholar of Shanghai Jiaotong University and Summer Undergraduate Research Fellowship (Purdue University)

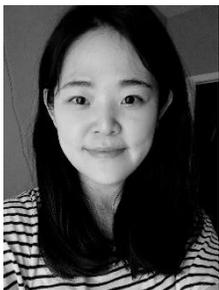
**Songan Zhang** received B.S. degree and M.S. degree in automotive engineering from Tsinghua University in 2013 and 2016 respectively. She is currently pursuing the Ph.D degree in mechanical engineering at University of Michigan, Ann Arbor.

Since 2016, she has been a graduate student at University of Michigan, Ann Arbor and her research interests include accelerated evaluation of autonomous vehicle and model-based reinforcement learning for autonomous vehicle decision making.

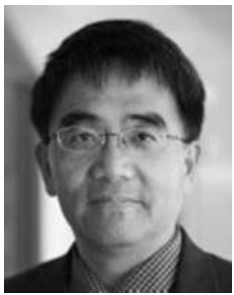
**Huei Peng** received the Ph.D. degree from the University of California, Berkeley, CA, USA, in 1992. He is currently a Professor with the Department of Mechanical Engineering, University of Michigan, Ann Arbor. He is currently the U.S. Director of the Clean Energy Research Center—Clean Vehicle Consortium, which supports 29 research projects related to the development and analysis of clean vehicles in the U.S. and in China. He also leads an education project funded by the Department of Energy to develop ten undergraduate and graduate courses, including three laboratory courses focusing on transportation electrification. He serves as the Director of the University of Michigan Mobility Transformation Center, a center that studies connected and autonomous vehicle technologies and promotes their deployment. He has more than 200 technical publications, including 85 in refereed journals and transactions. His research interests include adaptive control and optimal control, with emphasis on their applications to vehicular and transportation systems. His current research focuses include design and control of electrified vehicles, and connected/automated vehicles.